\documentclass{article}

\usepackage{arxiv}
\usepackage[utf8]{inputenc} 
\usepackage[T1]{fontenc}

\title{An Analysis and Implementation of Seam Carving for Content-Aware Image Resizing}

\author{
  Francesco Tosoni \\
  L'EMbeDS, Sant'Anna School of Advanced Studies \\
  Pisa, Italy \\
  \texttt{Francesco.Tosoni@santannapisa.it}
}

\date{} 

\usepackage{amsmath}
\usepackage{amssymb}
\usepackage{graphicx}

\usepackage[vlined,ruled]{algorithm2e}
\SetKwInOut{Input}{input}
\SetKwInOut{Output}{output}
\SetKwComment{Comment}{}{}

\SetCommentSty{mycmtsty}

\usepackage{amsthm}

\newtheorem*{remark}{Remark}

\newcommand{\Emap}{e}
\newcommand{\Mmap}{M}
\newcommand{\abs}[1]{\left|#1\right|}

\usepackage{hyperref}
\usepackage{cleveref}
\crefname{figure}{Figure}{Figures}
\Crefname{figure}{Figure}{Figures}
\crefname{section}{Section}{Sections}
\Crefname{section}{Section}{Sections}
\crefname{algorithm}{Algorithm}{Algorithms}
\Crefname{algorithm}{Algorithm}{Algorithms}

\begin{document}

\maketitle 

\begin{abstract}
Seam carving is a classical content-aware image resizing operator that modifies the width or height of an image by repeatedly removing (or inserting) seams, i.e., 8-connected monotonic paths of pixels of locally minimal importance. Because seams bend around salient content rather than uniformly scaling or cropping it, the operator preserves vital image structures while discarding (or duplicating) low-energy regions. This article describes a C++ implementation of the operator that follows the original formulation of Avidan and Shamir (2007), including the optional forward-energy criterion subsequently introduced by Rubinstein, Shamir and Avidan (2008). The implementation supports image reduction, image enlargement via ordered seam insertion, multi-pass enlargement for large scale factors, a user-supplied weight mask for object protection and removal, along with dumping of energy maps and visualisation of seams. We detail the algorithm, its parameters and its computational complexity, discuss design choices with respect to the original descriptions, and illustrate the behaviour of the operator on natural images.
\end{abstract}

\keywords{seam carving \and content-aware image resizing \and image retargeting \and forward energy \and dynamic programming \and object removal}

\paragraph{Source Code}
The C++ source code and documentation for this algorithm are publicly available at \url{https://github.com/ftosoni/seam-carving}. The implementation depends solely on a C++17 compiler and, optionally, OpenMP for multi-threading; image input/output relies on the public-domain \texttt{stb\_image} headers. Compilation (with CMake) and usage instructions are provided in the repository's \texttt{README.txt} file.

\section{Introduction}

Standard image resizing operators change an image's dimensions without considering its visual content. Uniform scaling distorts every object by the same factor, whereas cropping can only discard peripheral pixels. Neither approach is satisfactory when an image must be retargeted to a display or layout with an aspect ratio different from the original. Seam carving, introduced by Avidan and Shamir~\cite{siggraph2007}, is a discrete operator that resizes an image while preserving its salient regions. The key idea is to define an energy function that measures the importance of each pixel and to remove or insert seams: connected paths of pixels of minimal cumulative energy that cross the image from side to side. Removing a vertical seam reduces the image width by one pixel. The overall shape of the image is preserved because exactly one pixel is removed from each row. Yet, the visual content is maintained because the seam passes through the least important pixels.

A limitation of the original energy criterion is that it removes the seam of least energy in the \emph{current} image, ignoring the energy that the removal \emph{introduces} into the resulting image by making previously non-adjacent pixels neighbours. Rubinstein, Shamir and Avidan~\cite{siggraph2008} addressed this issue with a forward-energy criterion. At each step, this criterion removes the seam that introduces the least energy to the retargeted image. The forward criterion better protects straight and diagonal structures and reduces the staircase artefacts produced by the original backwards-energy formulation.

This article documents an implementation that supports both approaches, along with image enlargement via ordered seam insertion and a user-supplied weight mask for object protection and removal. \Cref{sec:operator} recalls the definition of a seam and the backward-energy dynamic programming framework. \Cref{sec:forward} describes the forward-energy criterion. \Cref{sec:reduction,sec:enlargement} describe the reduction and enlargement procedures. \Cref{sec:mask} details the masking strategy. \Cref{sec:implementation} provides implementation details, including parallelisation and a complexity analysis. \Cref{sec:examples} presents illustrative examples, and \Cref{sec:limitations} discusses the algorithm limitations.

\section{The Seam Carving Operator}
\label{sec:operator}

\subsection{Seams}

Let $I$ be an image of size $n\times m$ ($n$ rows and $m$ columns). The pixel dimensions of specific images quoted in the figures and examples follow the conventional width$\times$height order, i.e.\ $m\times n$. Following Avidan and Shamir~\cite{siggraph2007}, a \emph{vertical seam} is an $8$-connected path of pixels containing exactly one pixel in each row:
$$
s^x = \{(x(i), i)\}_{i=1}^{n},
\qquad \text{such that } \forall i,\; \abs{x(i)-x(i-1)} \le 1,
$$
where $x:[1,\dots,n]\to[1,\dots,m]$ maps each row $i$ to a column $x(i)$. The constraint $\abs{x(i)-x(i-1)}\le 1$ forces the path to be connected and monotonic in the row index. A \emph{horizontal seam} is defined symmetrically, containing one pixel per column. Removing a vertical seam shifts the pixels in the right columns leftwards by one column, reducing the width by one while keeping the image rectangular, as illustrated in \Cref{fig:scheme}.

\subsection{Energy}
\label{sec:energy}

Seam selection is driven by an energy function $\Emap$ that assigns a cost to each pixel. Avidan and Shamir define the gradient-magnitude energy.
$$
e_1(I) = \abs{\frac{\partial I}{\partial x}} + \abs{\frac{\partial I}{\partial y}},
$$
which is the $L^1$ norm of the image gradient. They also report having tested the $L^2$ norm of the gradient~\cite[Section~3.2]{siggraph2007}. Our implementation uses the $L^2$ variant
$$
\Emap(i,j) = \sqrt{\,(\partial_x I)^2 + (\partial_y I)^2\,},
$$
where the partial derivatives are estimated via central differences. For a single intensity channel, this is expressed as
$$
\partial_x I(i,j) = I(i,j{+}1) - I(i,j{-}1),
\qquad
\partial_y I(i,j) = I(i{+}1,j) - I(i{-}1,j),
$$
where out-of-range coordinates wrap around periodically. By default, the derivatives are computed on the luminance obtained from the BT.601 weights,
$$
Y = 0.299\,R + 0.587\,G + 0.114\,B,
$$
which approximates the perceptual response of the human eye. Optionally (\texttt{-{}-no-luma}), the energy is computed directly from the RGB channels as $\Emap(i,j)=\sqrt{\sum_{c}\big((\partial_x I_c)^2+(\partial_y I_c)^2\big)}$, summing the squared channel gradients before extracting the square root.

\begin{remark}
The missing $1/2$ scale factor in the standard central-difference formulation (compared with a forward difference) and the handling of periodic boundaries do not alter which seam is optimal: only the \emph{relative} energy of the pixels matters when minimising the cumulative cost.
\end{remark}

\subsection{Optimal Seam by Dynamic Programming}
\label{sec:dp}

\begin{figure}[tbp]
\centering 
\includegraphics[width=.9\linewidth]{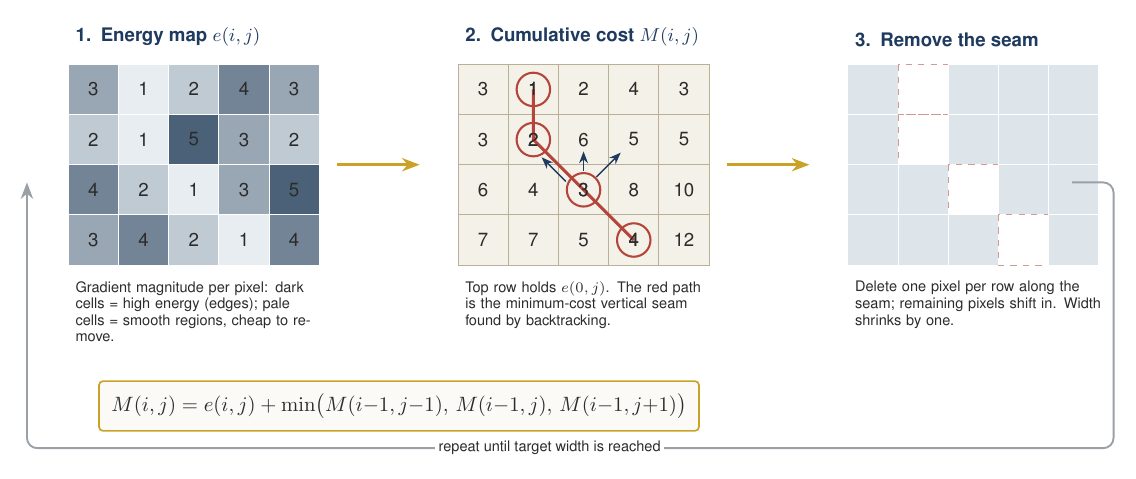}
\caption{One iteration of width reduction with backward energy on a $5\times4$ toy image. (1)~The energy $e(i,j)$ is the gradient magnitude (dark cells indicate high-energy edges, pale cells are smooth and cheap to remove). (2)~The cumulative cost $M(i,j)$ is filled from the top row downwards using the recurrence shown; the minimum value of the last row is the cost of the optimal seam, which is recovered via backtracking (red path). (3)~The seam is deleted, one pixel per row, and the remaining pixels shift inwards, reducing the width by one. The loop repeats until the target width is reached.}
\label{fig:scheme}
\end{figure}

The cost of a seam is the sum of the energy of its constituent pixels, $E(s)=\sum_{i=1}^{n}\Emap(s_i)$, and the optimal seam is the one that minimises this cost. It is found via dynamic programming~\cite[\S15]{cormen-book}. We first build the cumulative minimum energy map $\Mmap$ by scanning the image from the second row to the last:
$$
\Mmap(i,j) = \Emap(i,j) +
\min\big(\Mmap(i{-}1,j{-}1),\, \Mmap(i{-}1,j),\, \Mmap(i{-}1,j{+}1)\big),
$$
with $\Mmap(1,j)=\Emap(1,j)$, where out-of-range neighbours are omitted at the left and right borders. The minimum value in the last row of $\Mmap$ corresponds to the cost of the optimal seam. The seam itself is recovered by backtracking from that minimum to the top row, moving at each step to the neighbour among $\{j{-}1,j,j{+}1\}$ that realised the minimum. Horizontal seams are obtained by applying the same procedure to the transposed image. \Cref{fig:scheme} illustrates these steps on a small toy image (the energy map, the cumulative cost $\Mmap$, and the seam recovered by backtracking), together with the pixel removal that completes one reduction iteration.

\section{Forward Energy}
\label{sec:forward}

\begin{figure}[t]
\centering
\includegraphics[width=.8\linewidth]{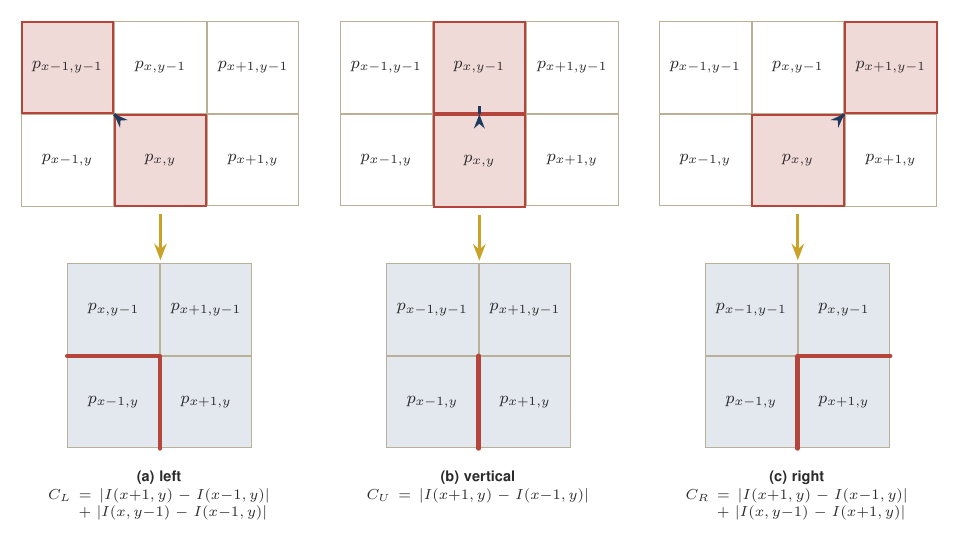}
\caption{The three forward-energy step costs for a vertical seam passing through a pixel~\cite[Figure~7]{siggraph2008}. For a left step (a), a vertical step (b), and a right step (c), the seam removes the highlighted pixel and its predecessor (top grid); the surviving neighbours then regroup (bottom grid), and the newly created pixel edge is drawn in red. The associated costs $C_L$, $C_U$, or $C_R$ count exactly these new edges. Pixels are indexed in image coordinates $p_{x,y}$, with $x$ the column and $y$ the row, as in the implementation; this corresponds to the $C_\bullet(i,j)$ of the text under the substitution $j\mapsto x$, $i\mapsto y$.}
\label{fig:forward-edges}
\end{figure}

The backward energy formulation described in \Cref{sec:energy} ranks seams based on the energy they \emph{remove}. As observed by Rubinstein \emph{et al.}~\cite{siggraph2008}, this ignores the energy that the removal process \emph{introduces}. When a seam pixel is deleted, its former left and right neighbours become adjacent, creating a new pixel edge. The forward-energy criterion charges each step with the cost of the new edges it creates, ensuring that the algorithm prefers seams that leave the resulting image as smooth as possible.

Concretely, for a vertical seam passing through pixel $(i,j)$, there are three possible predecessors, and each is assigned a forward cost~\cite[Section~5.1]{siggraph2008}:
$$
\begin{aligned}
C_L(i,j) &= \abs{I(i,j{+}1) - I(i,j{-}1)} + \abs{I(i{-}1,j) - I(i,j{-}1)},\\
C_U(i,j) &= \abs{I(i,j{+}1) - I(i,j{-}1)},\\
C_R(i,j) &= \abs{I(i,j{+}1) - I(i,j{-}1)} + \abs{I(i{-}1,j) - I(i,j{+}1)}.
\end{aligned}
$$

The term $C_U$ represents the cost of the horizontal edge created between the left and right neighbours of the removed pixel, whereas $C_L$ and $C_R$ add the cost of the new vertical edge created when the seam steps in from the left or right diagonal (see \Cref{fig:forward-edges}). The cumulative cost map is then defined as
$$
\Mmap(i,j) = P(i,j) + \min
\begin{cases}
\Mmap(i{-}1,j{-}1) + C_L(i,j),\\
\Mmap(i{-}1,j) + C_U(i,j),\\
\Mmap(i{-}1,j{+}1) + C_R(i,j),
\end{cases}
$$
where $P(i,j)$ is an optional per-pixel energy term that can be supplied on top of the forward cost, such as a saliency value, the output of a face detector, or a user-defined weight (\Cref{sec:mask}). The first row is initialised with $\Mmap(1,j)=P(1,j)$. As with backward energy, the differences $\abs{\cdot}$ are evaluated on the luminance (or, in RGB mode, as the corresponding norm of the channel differences), and out-of-range coordinates wrap around. Backtracking recomputes $C_L$, $C_U$, and $C_R$ along the path so that the descent follows the same costs used to build $\Mmap$.

\section{Image Reduction}
\label{sec:reduction}

To reduce the width from $m$ to $m'<m$, the operator sequentially removes $m-m'$ vertical seams, recomputing the energy and the cumulative cost map after each removal (\Cref{alg:reduce}); \Cref{fig:scheme} depicts one such iteration. For cache-efficiency reasons, height reduction is performed by transposing the image, removing vertical seams from the transpose, and transposing back, meaning that a single vertical-seam routine serves both directions. When both dimensions must change, our implementation processes width before height.

The original method also defines an \emph{optimal} order for interleaving horizontal and vertical seam removals, obtained via a transport map computed by dynamic programming~\cite[Section~4.2]{siggraph2007}. The transport map identifies, for every intermediate target size, the least total energy that must be removed to reach it, and backtracking through it yields the optimal sequence of horizontal and vertical removals. The present implementation uses the simpler fixed schedule described above, removing all width seams and then all height seams. This approach is faster and uses less memory, but it does not guarantee the globally optimal removal order when both dimensions change.

\begin{algorithm}[!htbp]
\caption{Width reduction by seam removal}
\label{alg:reduce}
\DontPrintSemicolon
\Input{image $I$ of width $m$, target width $m'$, energy mode}
\Output{image of width $m'$}
\While{$m > m'$}{
  compute the energy map $\Emap$ of $I$\;
  build the cumulative cost map $\Mmap$ (backward or forward)\;
  $j^\star \leftarrow \arg\min_j \Mmap(n, j)$ \Comment*{end of optimal seam}
  backtrack from $(n, j^\star)$ to obtain the seam $s$\;
  remove $s$ from $I$ (and from the weight mask, if present)\;
  $m \leftarrow m - 1$\;
}
\end{algorithm}

\section{Image Enlargement}
\label{sec:enlargement}

Enlarging an image by inserting seams cannot simply repeat the process of finding the optimal seam and duplicating it; doing so would repeatedly select the same minimal seam, producing a stretching artefact along a single path. Following Avidan and Shamir~\cite[Section~4.3]{siggraph2007}, to enlarge the width by $k$ pixels, we first compute the $k$ seams that \emph{would} be removed in order, and then duplicate them in the original image. The first $k$ seams are obtained on a working copy, with each seam removed before the next is sought. An index map tracks the original column position of each pixel in the shrinking copy, allowing the seams to be translated back to the original coordinates. Each inserted pixel is computed as the average of the seam pixel and an adjacent neighbour.

Duplicating all the seams of an image at once is equivalent to uniform scaling, meaning that inserting several seams comparable to the current width loses the content-aware advantage. To enlarge an image by a large factor, the process is therefore split into several passes. Each pass inserts at most half as many seams as the current width (enlarging by at most 50\% per pass), and the procedure repeats until the target width is achieved. Height enlargement reduces to the width case via transposition. \Cref{alg:insert} summarises one enlargement pass.

\begin{algorithm}[!htbp]
\caption{Width enlargement by ordered seam insertion (one pass)}
\label{alg:insert}
\DontPrintSemicolon
\Input{image $I$ of width $m$, number of seams $k \le \lfloor m/2 \rfloor$}
\Output{image of width $m+k$}
$C \leftarrow$ working copy of $I$;\quad initialise index map to identity\;
\For{$t \leftarrow 1$ \KwTo $k$}{
  find the optimal seam $s$ on $C$\;
  record $s$ in original coordinates via the index map\;
  remove $s$ from $C$ and update the index map\;
}
\ForEach{row}{
  sort the recorded seam columns and reinsert them from left to right,\;
  setting each new pixel to the average of the seam pixel and a neighbour\;
}
\end{algorithm}

\section{Masking: Object Protection and Removal}
\label{sec:mask}

The operator accepts an optional weight mask of the same size as the input, which adds a per-pixel term to the cumulative cost: the term $P(i,j)$ in \Cref{sec:forward}; for backward energy, it is added to $\Mmap$ in the same manner. A large positive weight forces the seams to avoid a region, protecting it, whereas a large negative weight attracts the seams, ensuring that the region is carved away first. This strategy achieves the object protection and object removal applications described in the original paper~\cite[Section~4.6]{siggraph2007} without requiring a dedicated removal loop. The masked object is eliminated as a side effect of reducing the image width beyond the object's horizontal extent, and the original dimensions can be restored afterwards via seam insertion.

In the command-line tool, the mask is read from an image file. In a colour mask, the weight of a pixel is proportional to the difference $G-R$ between its green and red channels, where green marks protection and red marks removal. We note that the original paper adopts the opposite colour coding, green for removal and red for protection~\cite[Figure~11]{siggraph2007}; we instead follow the more conventional association of red with removal and green with protection. In a greyscale mask, sufficiently bright pixels protect content, sufficiently dark pixels mark it for removal, and intermediate grey values leave the cost unchanged. To remove an object successfully, the image must be carved by at least the object's width to ensure that enough seams pass through the marked region.

\section{Implementation and Complexity}
\label{sec:implementation}

\subsection{Complexity}

Building the energy map and the cumulative cost map each requires $O(n\,m)$ operations for an $n\times m$ image, and recovering a seam by backtracking costs $O(n)$. Removing a single vertical seam, including the subsequent reconstruction of the image, costs $O(n\,m)$. Reducing the width by $c$ seams therefore costs $O(c\,n\,m)$, demonstrating that the operator scales linearly with both the number of pixels and the number of seams removed. Enlargement shares the same asymptotic cost: a single pass that first inserts $k\le m$ seams, then extracts $k$ seams from a working copy in $O(k\,n\,m)$ time, and finally rebuilds the image once. Memory usage is $O(n\,m)$ because the storage required for the image, the energy map, the cumulative cost map, the optional weight mask, and the index map (used during enlargement) is proportional to the total number of pixels.

\subsection{Parallelisation}

The implementation is written in C++17 and is optionally parallelised using OpenMP. Parallelism is applied where operations are independent and free of inter-row synchronisation. This includes generating the energy map, whose rows are mutually independent and distributed across the available threads; the seam-removal and seam-insertion stages, which reconstruct the image row by row; and the transpose operation used for horizontal seams. By contrast, the cumulative-cost recurrence relation is evaluated serially. Although the entries of a single row are mutually independent, each row depends directly on the results of the previous one. Consequently, a parallel inner loop would impose a thread-team barrier after every row. The synchronisation overhead of this barrier exceeds the arithmetic intensity of a row even for large images; hence, a serial scan is consistently faster, and parallelising the recurrence is not worthwhile. The number of threads is configurable, defaulting to the maximum hardware concurrency, and the final output is independent of the thread count.

\subsection{Parameters and Usage}

The operator exposes the following parameters:
\begin{description}
\item[target width and height] the dimensions of the output image; either can be smaller (reduction) or larger (enlargement) than the input.
\item[energy criterion] backward energy (default) or forward energy (\texttt{-{}-forward}).
\item[luminance] determines whether the energy is computed on the BT.601 luminance (default) or directly on the RGB channels (\texttt{-{}-no-luma}).
\item[mask] an optional weight image used for object protection and removal.
\item[threads] the number of OpenMP threads to allocate.
\end{description}

\section{Examples}
\label{sec:examples}

\begin{figure}[!htbp]
\begin{center}
\includegraphics[width=0.31\linewidth]{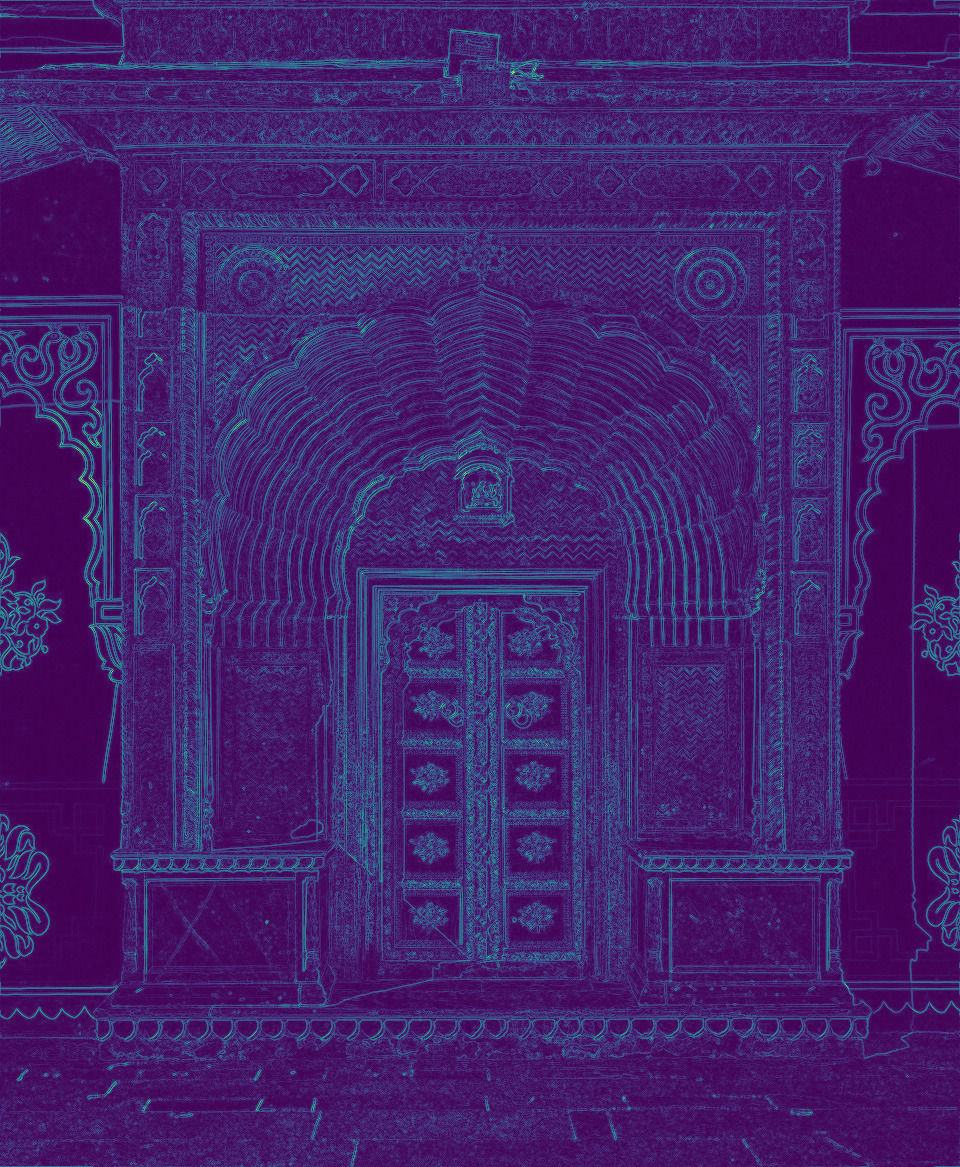}\hfill
\includegraphics[width=0.31\linewidth]{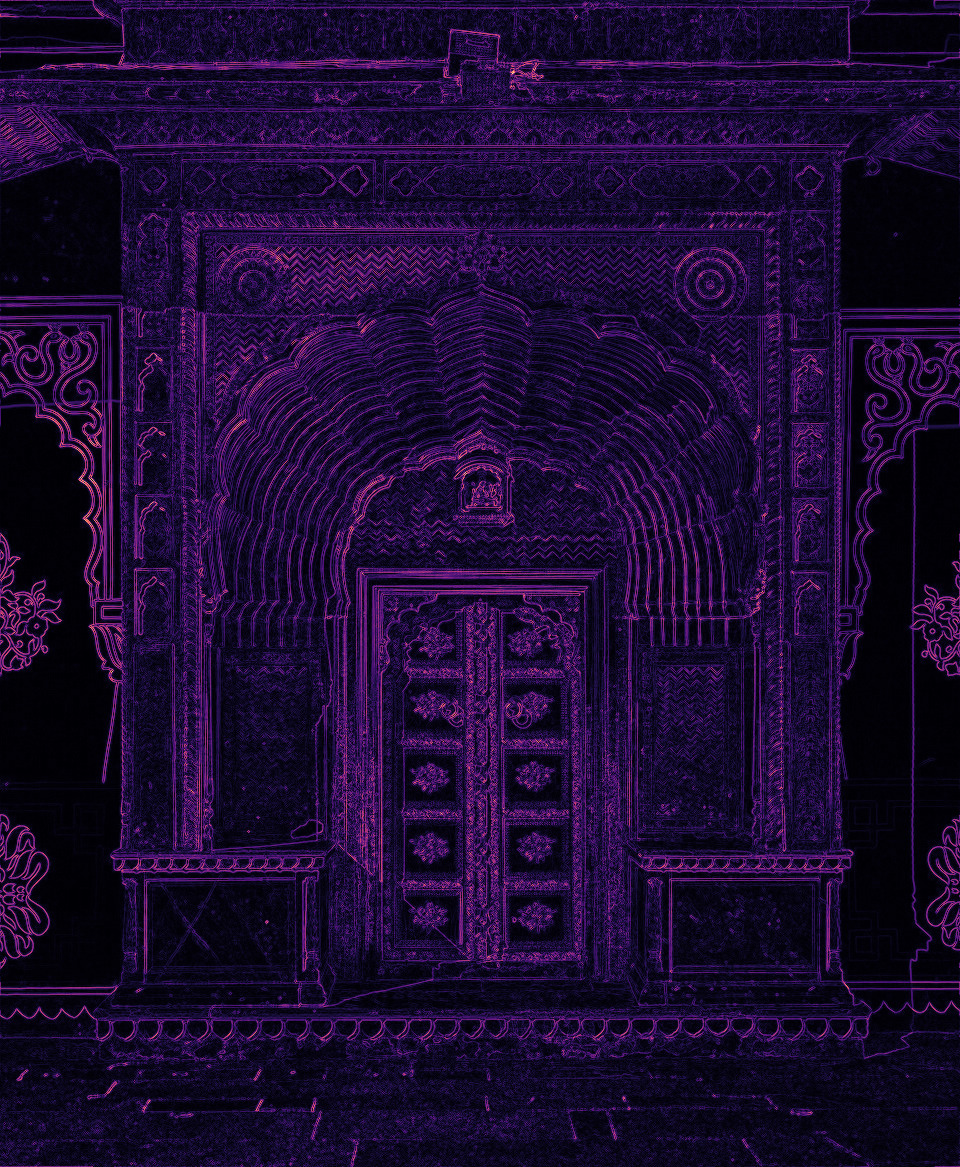}\hfill
\includegraphics[width=0.31\linewidth]{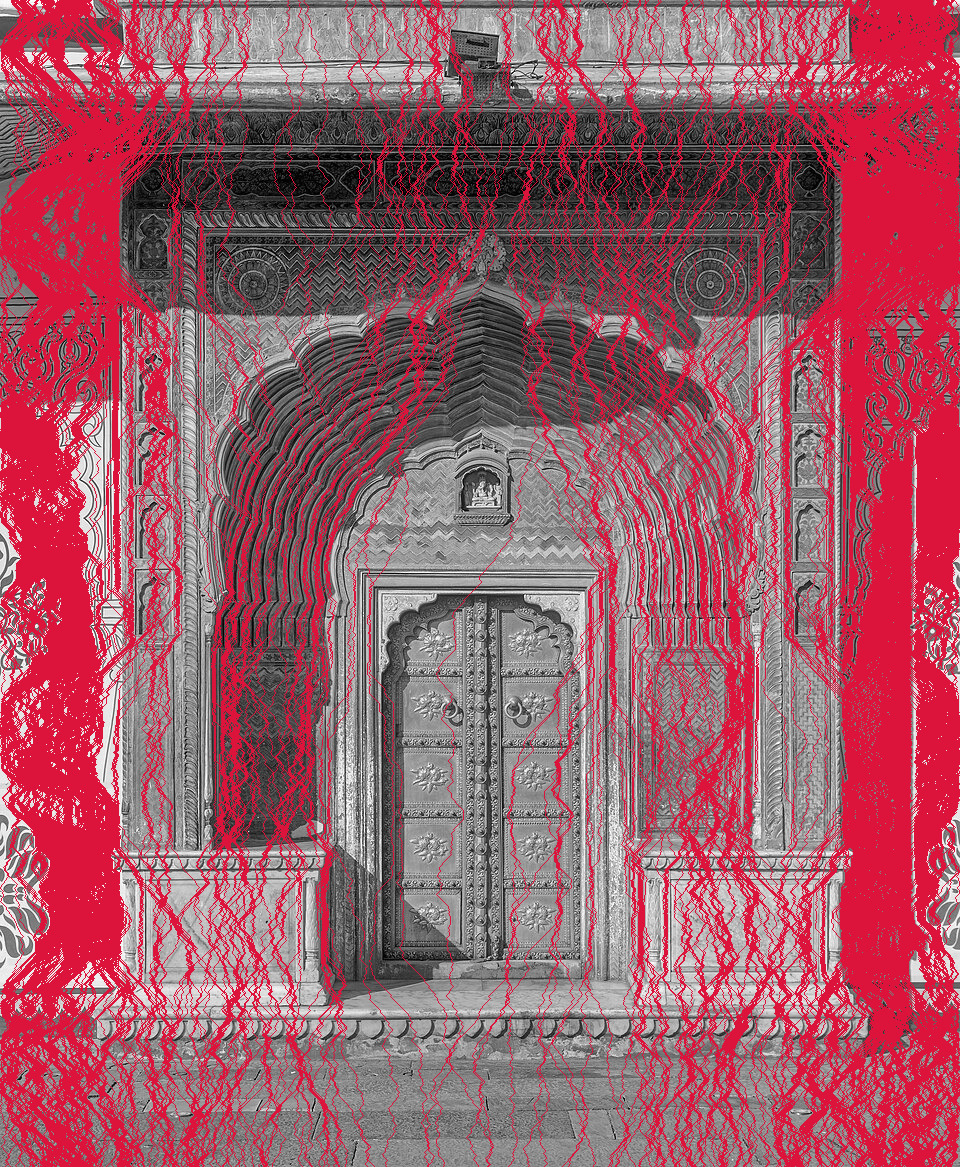}
\caption{Energy and selected seams for the first example (the portal of \Cref{fig:reduce}). Left: the backward dual-gradient energy map $\Emap$, rendered with the perceptually uniform \emph{viridis} colour map, where dark denotes low energy and bright denotes high energy. Middle: the same energy map rendered with the \emph{magma} colour map (\texttt{-{}-viz-colourmap magma}), an alternative perceptually uniform palette. Right: the $320$ vertical seams removed to reduce the width from $960$ to $640$ with backward energy, drawn in crimson over a greyscale copy of the input. The seams concentrate in the smooth walls and avoid the doorway, which is why its proportions are preserved under reduction.}
\label{fig:energy}
\end{center}
\end{figure}

Before presenting the resized results, \Cref{fig:energy} visualises the two quantities that drive the operator on the first example: the backward energy map and the seams selected for removal. The energy is largest along the carved doorway and its ornamental relief and smallest across the plain flanking walls; accordingly, the seams removed to reduce the width gather in those smooth regions and bend around the salient doorway.

\begin{figure}[!htbp]
\begin{center}
\includegraphics[width=0.247\linewidth]{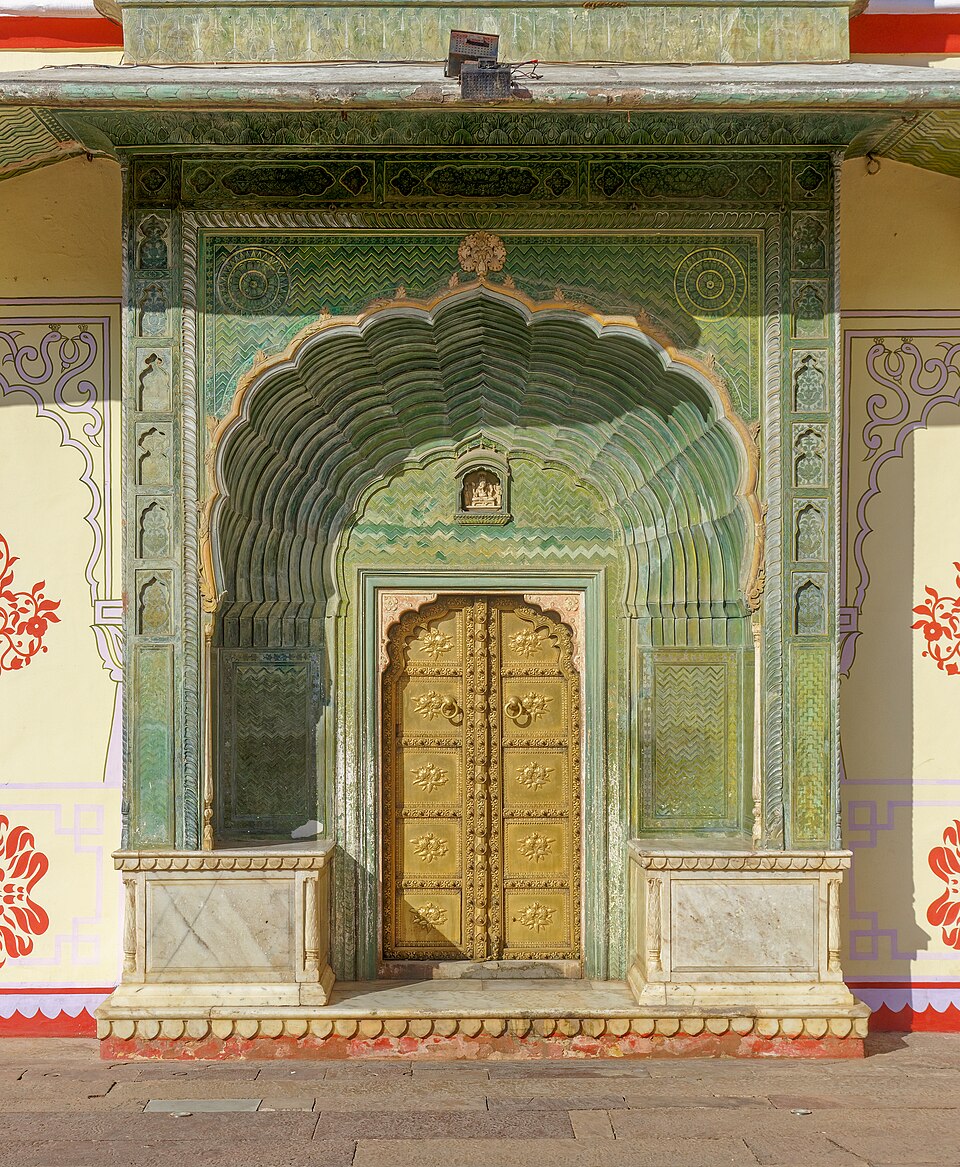}\hfill
\includegraphics[width=0.329\linewidth]{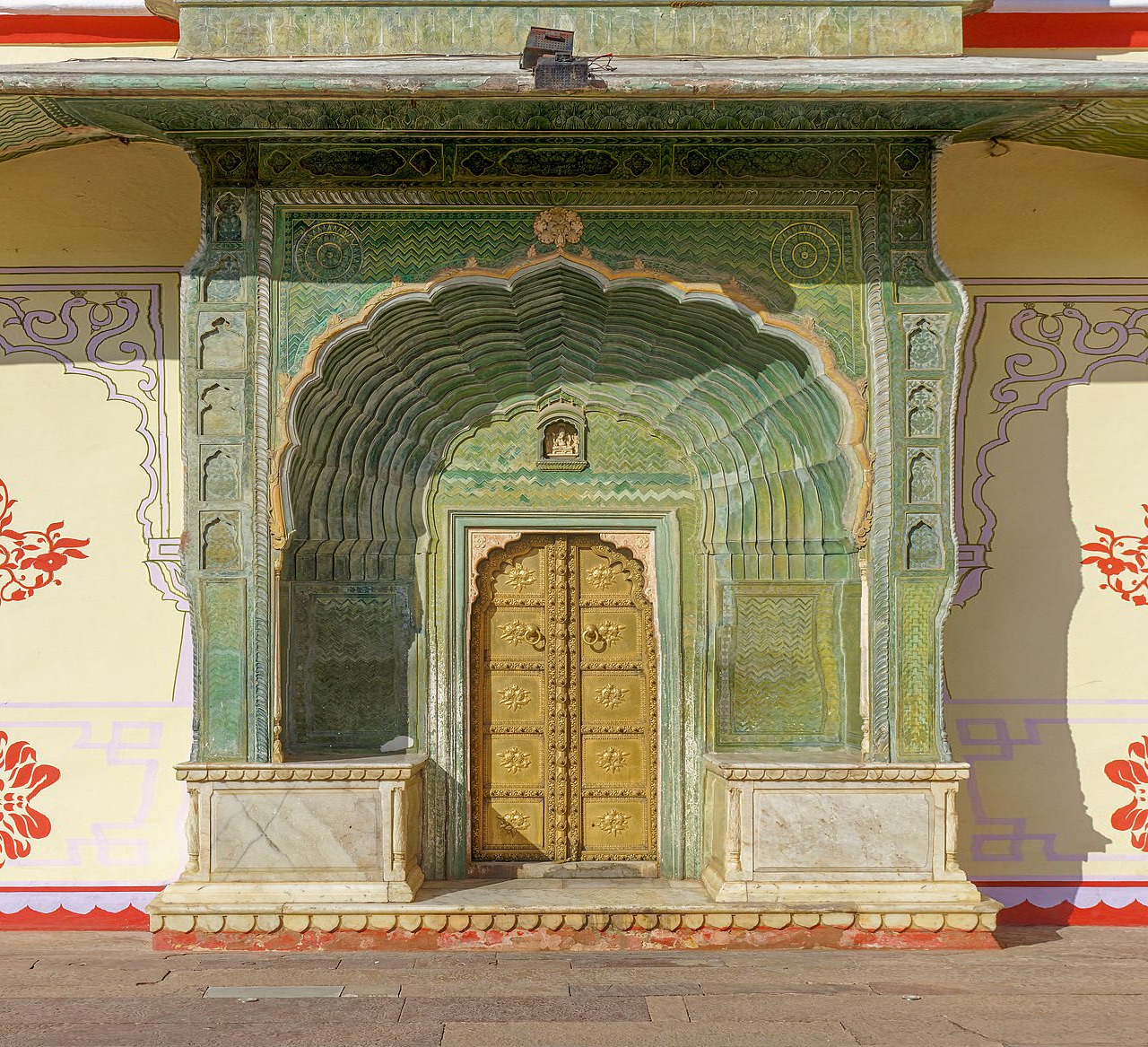}\hfill
\includegraphics[width=0.164\linewidth]{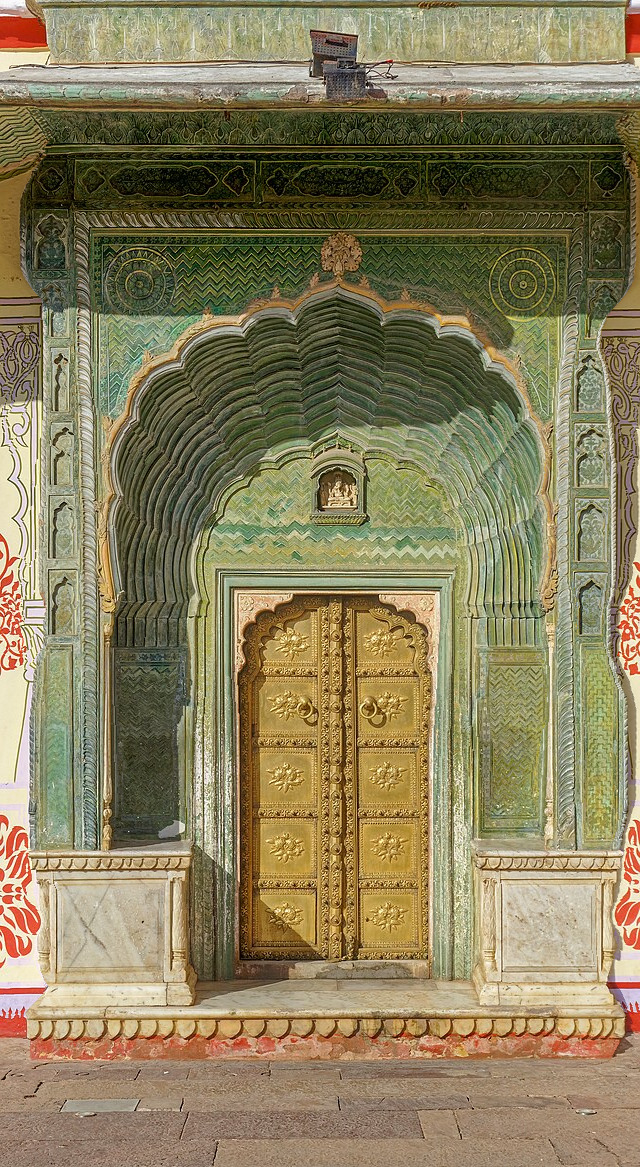}\hfill
\includegraphics[width=0.164\linewidth]{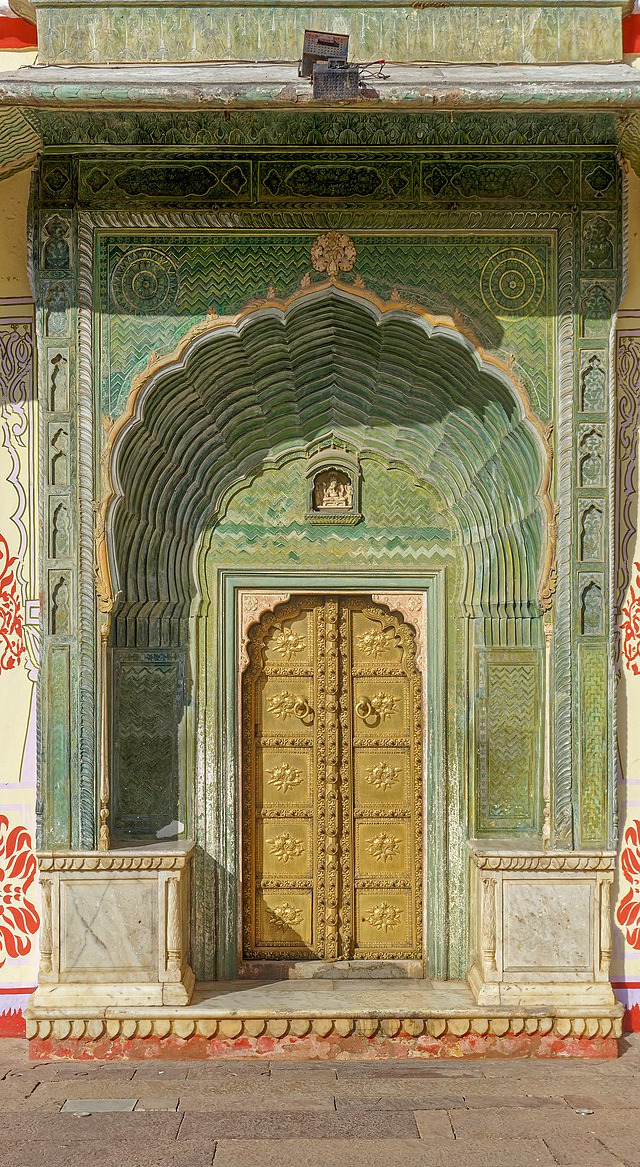}
\caption{Content-aware resizing. From left: original ($960\times1167$); enlarged to width $1280$ via ordered seam insertion; width reduced to $640$ with backward energy; width reduced to $640$ with forward energy. The four panels share a common height, so the enlarged panel appears wider and the reduced ones narrower. The doorway and its arch maintain their proportions while the plain flanking walls absorb the resizing.}
\label{fig:reduce}
\end{center}
\end{figure}

\Cref{fig:reduce} displays the width reduction of a natural image using both the backward and forward criteria, alongside content-aware enlargement of the same image. In all cases, the ornate doorway maintains its original proportions while the plain flanking walls absorb the change in width. In contrast, uniform scaling would compress or stretch the doorway horizontally.

\begin{figure}[!htbp]
\begin{center}
\includegraphics[width=0.44\linewidth]{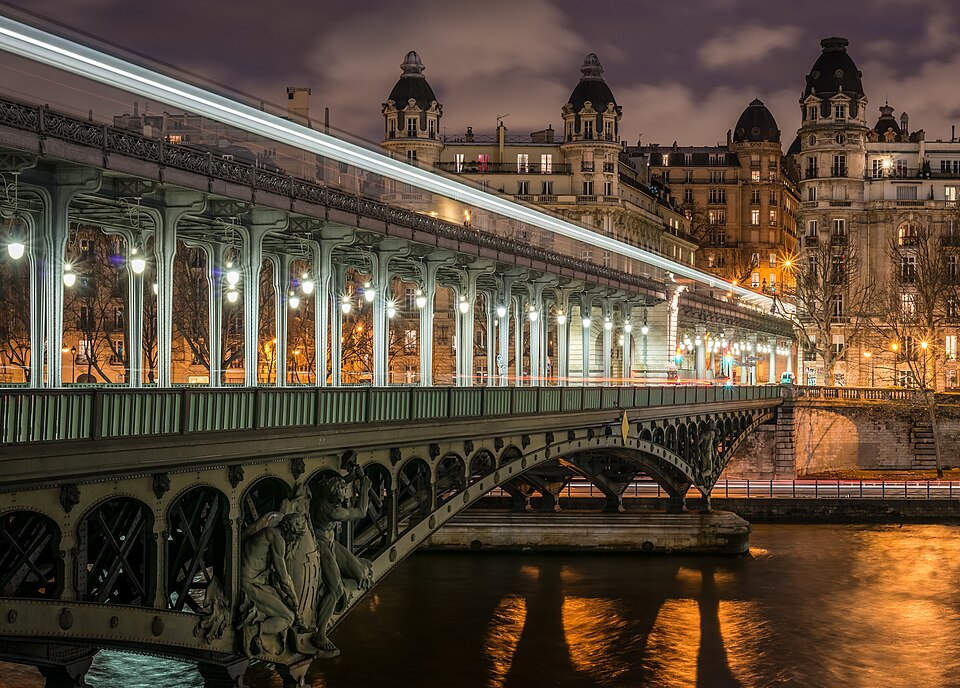}\hfill
\includegraphics[width=0.257\linewidth]{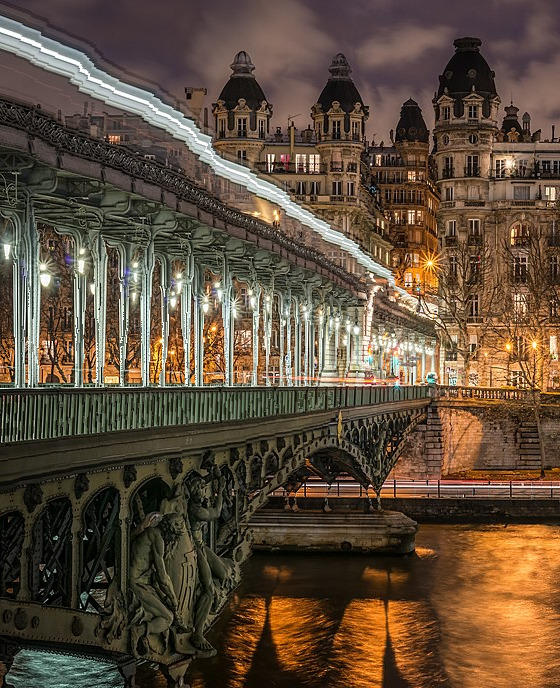}\hfill
\includegraphics[width=0.257\linewidth]{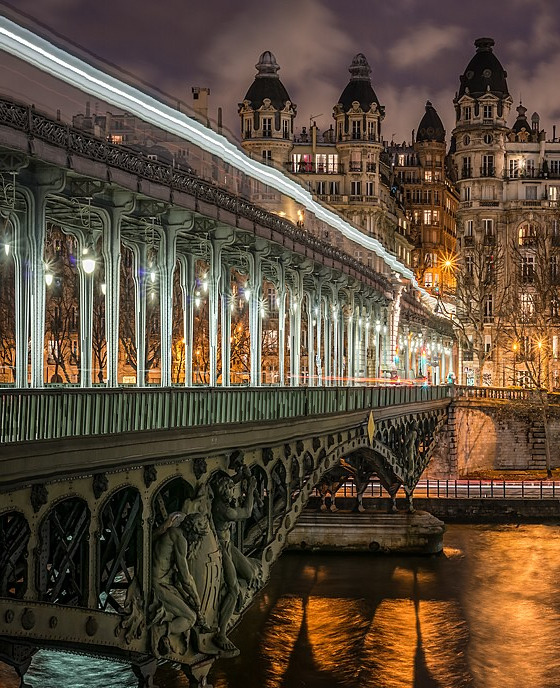}
\caption{Backward versus forward energy on an image dominated by straight architectural lines. Left: original ($960\times688$). Middle: width reduced to $560$ with backward energy. Right: width reduced to $560$ with forward energy. The three panels share a common height, so the carved ones appear narrower.}
\label{fig:forward}
\end{center}
\end{figure}

\Cref{fig:forward} compares the two criteria on an image dominated by straight architectural lines. The forward criterion preserves the regular colonnade of the viaduct and other linear architectural elements more effectively because it avoids seams whose removal would bridge pixels across these edges; backward energy warps these straight structures more visibly.

\begin{figure}[!htbp]
\begin{center}
\includegraphics[width=0.45\linewidth]{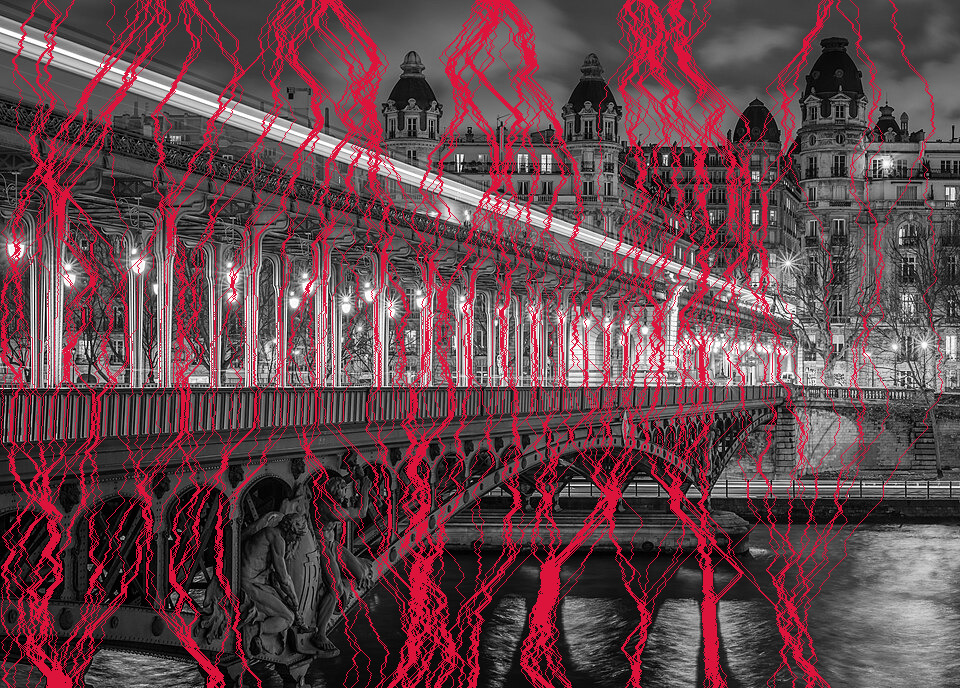}\hfill
\includegraphics[width=0.45\linewidth]{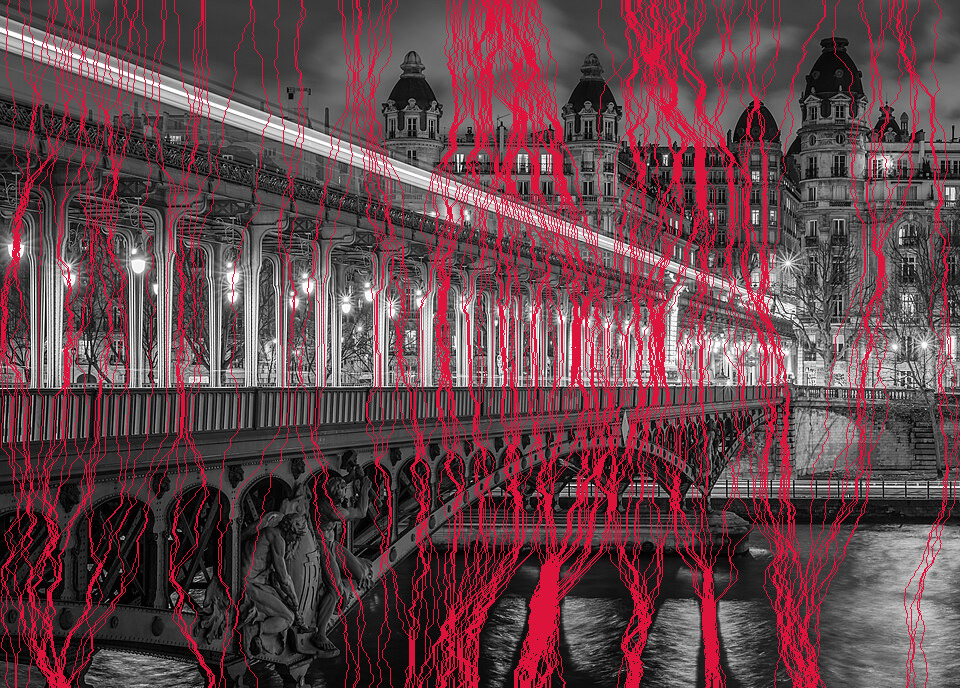}
\caption{Seam selection underlying \Cref{fig:forward}, shown for the first $160$ vertical seams each criterion would remove, drawn in crimson over a greyscale copy of the input. Left: backward energy. Right: forward energy.}
\label{fig:forwardseams}
\end{center}
\end{figure}

\Cref{fig:forwardseams} makes this mechanism visible by overlaying the seams that each criterion selects for \Cref{fig:forward}. The backward seams wander across the colonnade and the bridge railings. In contrast, the forward seams stay markedly straighter and gather in the smooth water and sky, which is precisely why they leave the architectural lines intact.

\begin{figure}[!htbp]
\begin{center}
\includegraphics[width=0.45\linewidth]{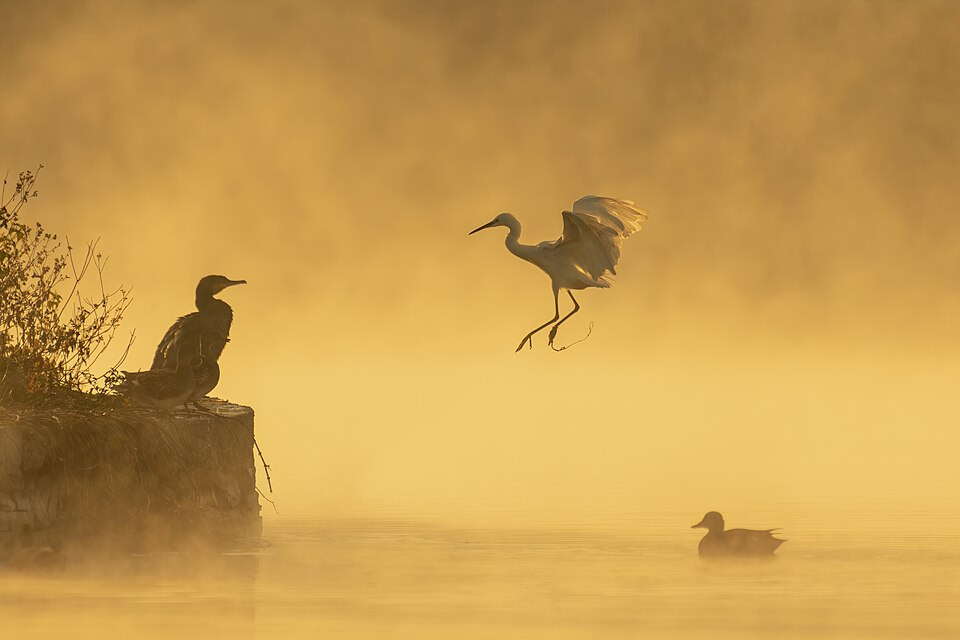}\hfill
\includegraphics[width=0.45\linewidth]{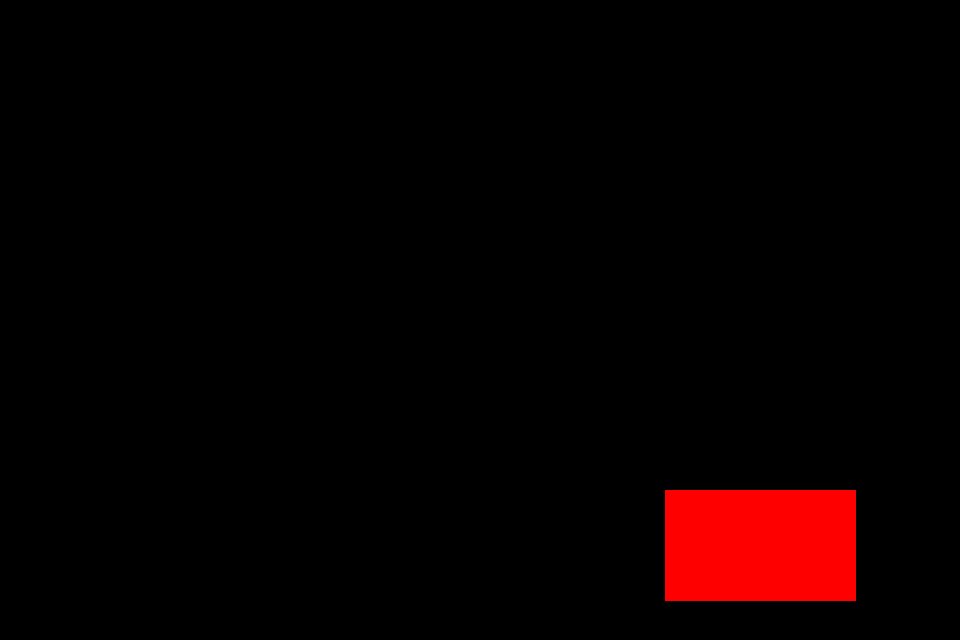}\\
\includegraphics[width=0.45\linewidth]{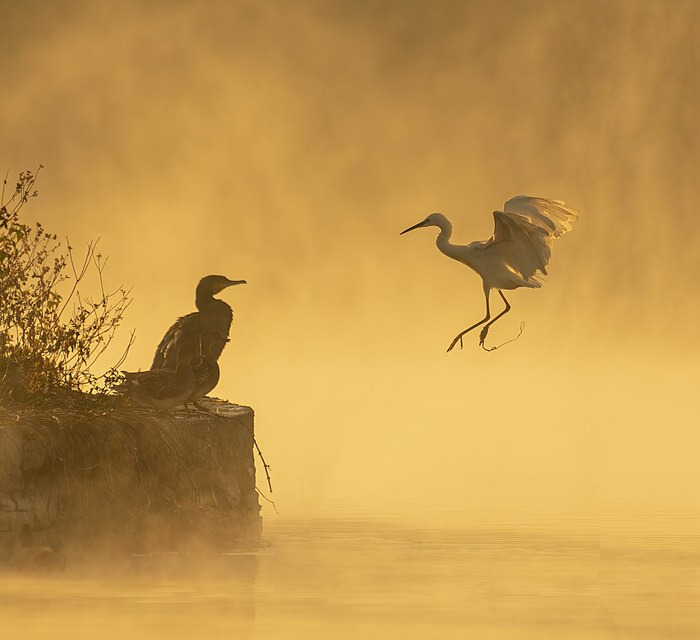}\hfill
\includegraphics[width=0.45\linewidth]{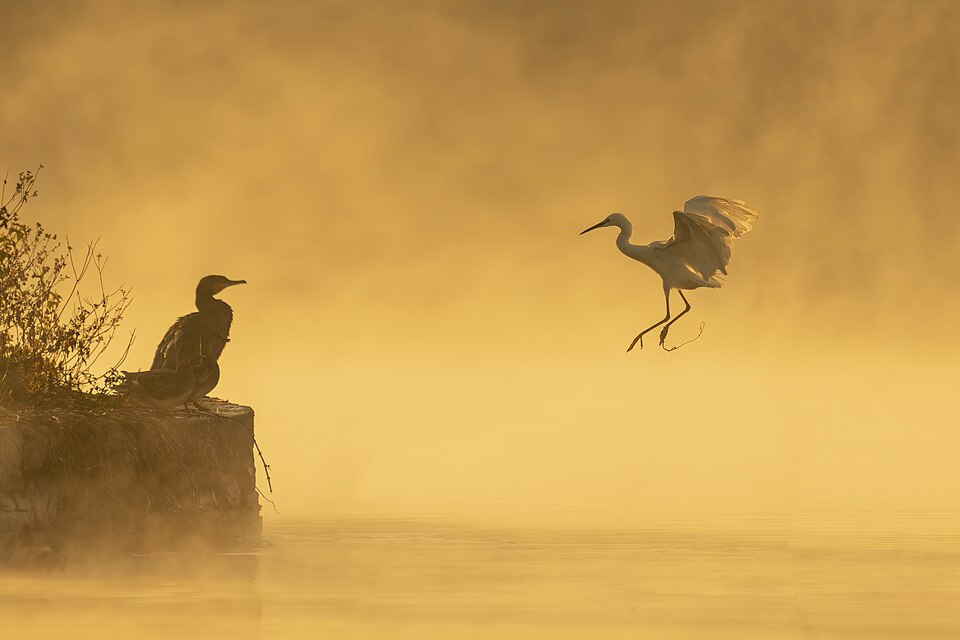}
\caption{Object removal with a weight mask. Top left: original. Top right: the removal mask (the marked region is attracted by negative weight). Bottom left: after carving the width to $700$, the duck is removed. Bottom right: the result, enlarged back to the original width of $960$, by seam insertion.}
\label{fig:removal}
\end{center}
\end{figure}

\Cref{fig:removal} illustrates object removal using a mask. The duck on the right is marked for removal; the image is carved until the duck is eliminated, then enlarged back to its original width via seam insertion. As a result, the output image matches the input's dimensions but no longer contains the duck element.

\begin{figure}[!htbp]
\begin{center}
\begin{minipage}[b]{0.32\linewidth}
\includegraphics[width=\linewidth]{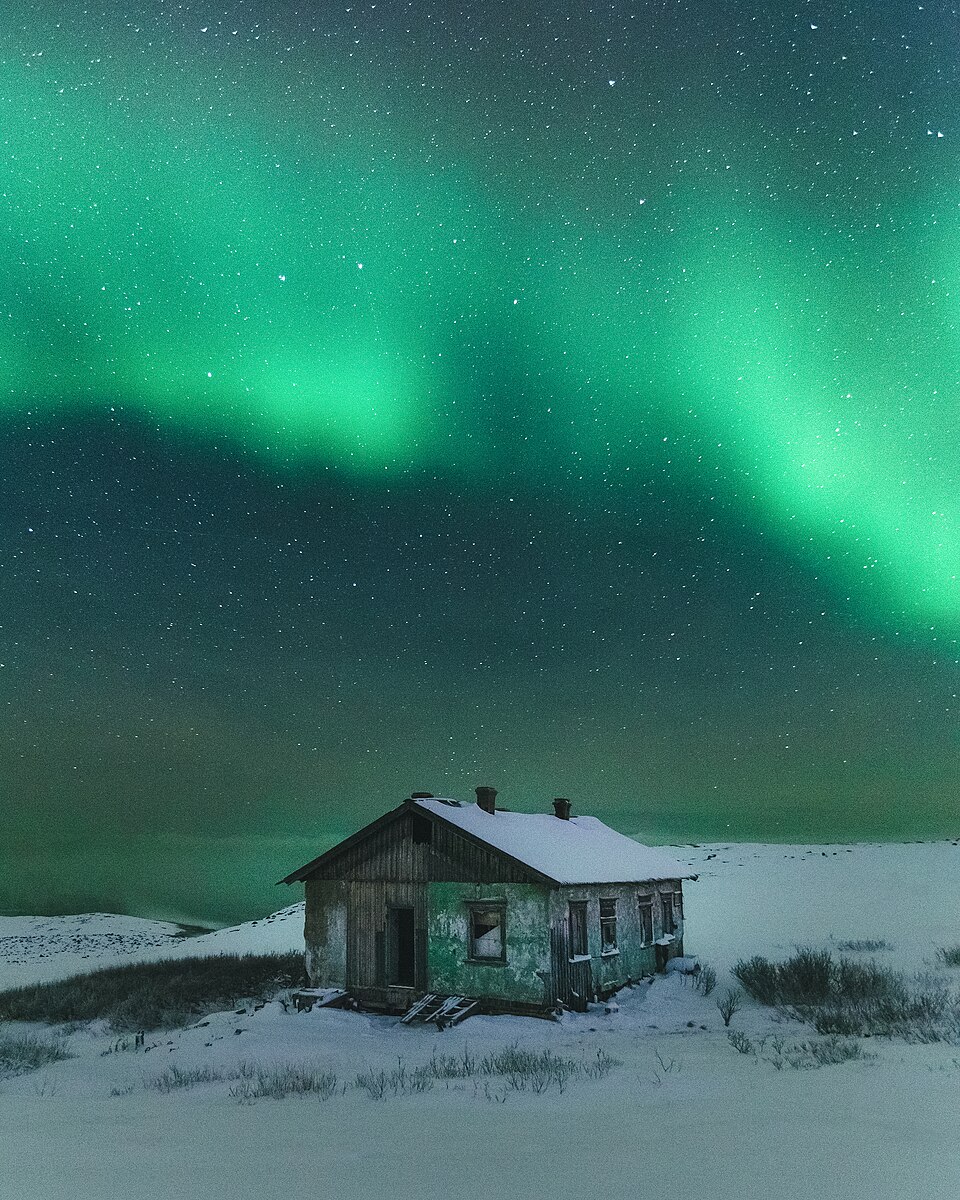}
\end{minipage}\hfill
\begin{minipage}[b]{0.32\linewidth}
\includegraphics[width=\linewidth]{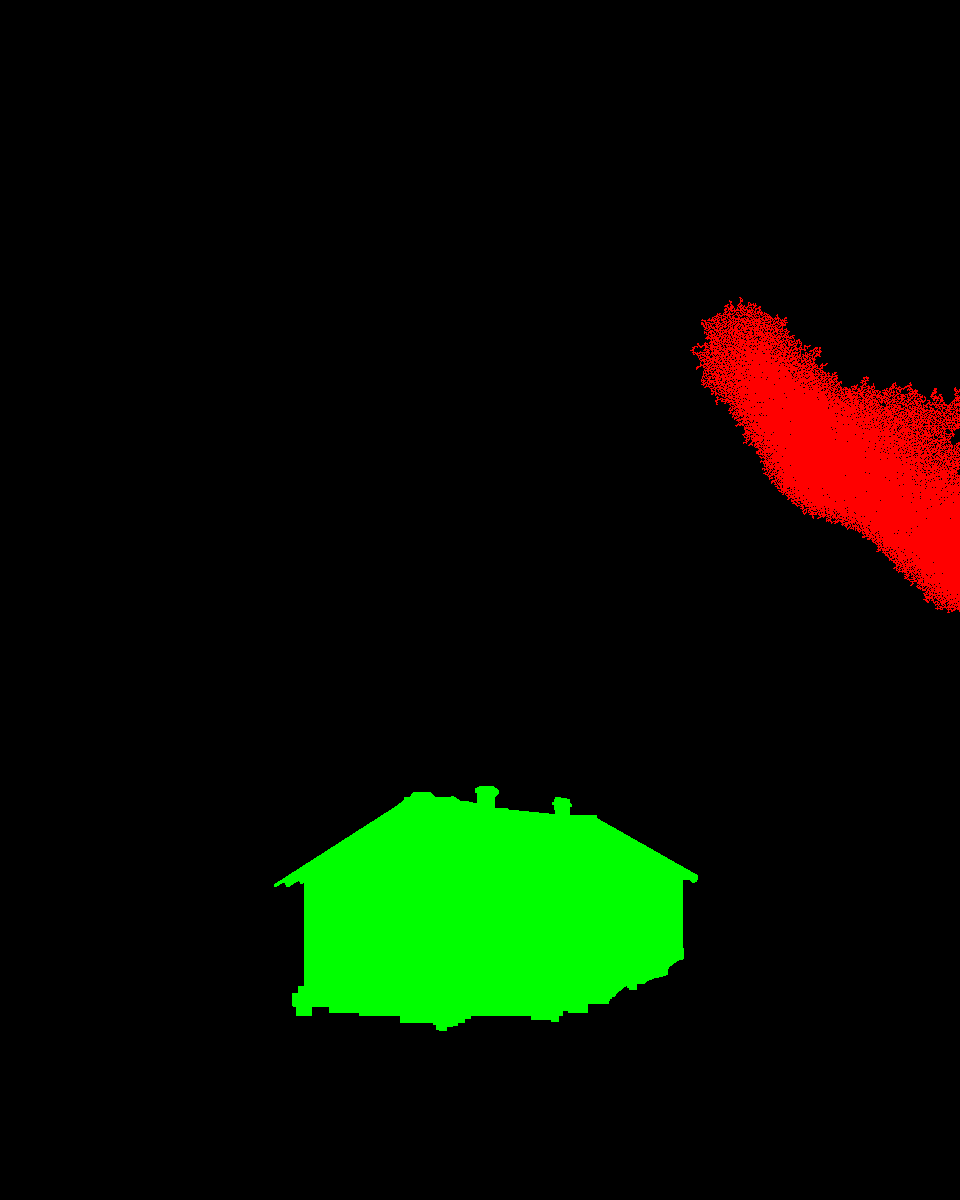}
\end{minipage}\hfill
\begin{minipage}[b]{0.32\linewidth}
\includegraphics[width=\linewidth]{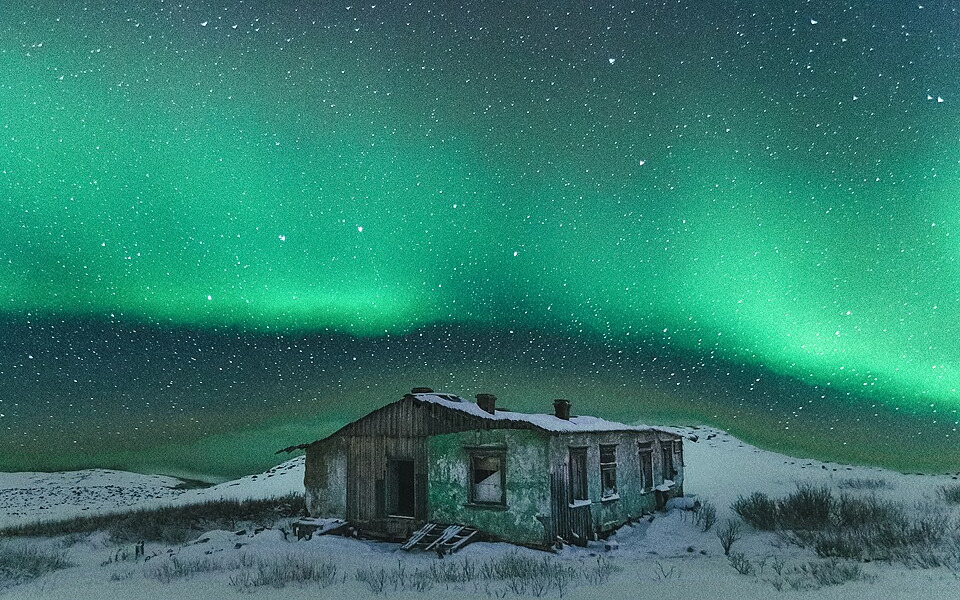}\\[2pt]
\includegraphics[width=\linewidth]{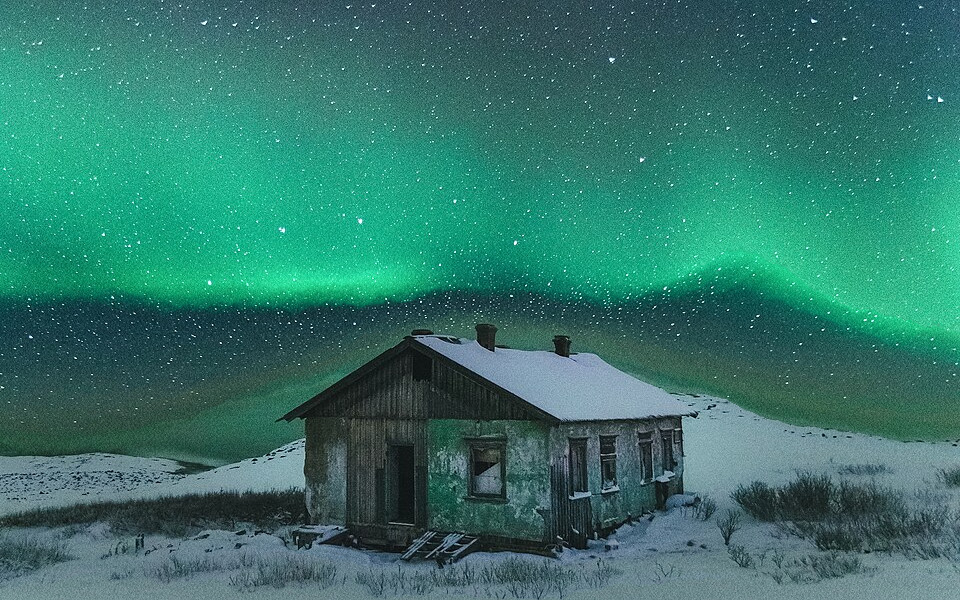}
\end{minipage}
\caption{Object protection and simultaneous removal with a weight mask. Left: original ($960\times1200$). Middle: the mask, where green protects the abandoned house and red marks a patch of aurora for removal. Right, top: height reduced to $600$ \emph{without} a mask, where the house is squeezed and distorted. Right, bottom: the same reduction \emph{with} the mask, where the house keeps its height and proportions (green) while the marked aurora patch is carved away first (red). The two stacked results share the width and height of one original column, so the vertical compression is read directly against the unaltered image.}
\label{fig:protect}
\end{center}
\end{figure}

The mask can also protect content, and protection and removal can be combined within a single mask. \Cref{fig:protect} depicts an image where a small, low-contrast house is squeezed and distorted by an unguided reduction, because the surrounding snow and sky are smooth, causing the seams to cut through the building. Painting the house green (positive weight) forces the seams to bypass it, allowing it to maintain its height and proportions while the smooth surroundings absorb the reduction. The same mask simultaneously paints a patch of aurora red (negative weight) to ensure it is carved away first. A single mask can therefore simultaneously preserve one region and remove another.

\section{Limitations}
\label{sec:limitations}

As noted in the original papers, the seam carving algorithm does not perform successfully on every image~\cite{siggraph2007,siggraph2008}. When an image is too densely packed with important content, no low-energy paths exist, and any chosen seam will inevitably cross salient structures. Similarly, when the spatial layout places important content across the entire width or height, seams cannot bypass it. In both scenarios, standard scaling or cropping may be preferable. Furthermore, while the forward criterion protects structure, it can do so at the expense of content; an object that could otherwise be resized without visible artefacts might be sacrificed to keep a highly textured background intact~\cite[Section~7]{siggraph2008}. Finally, the energy formulation used here is a low-level gradient measure. For images where structural or semantic importance is not well captured by image gradients, an external saliency map or explicit user weights via the mask are required to guide the operator.

\clearpage
\section*{Image Credits}

{\small
\includegraphics[height=2em]{images/portal_orig.jpg}
  Jakub Ha\l{}un (Wikimedia Commons),
  CC BY-SA 4.0\footnote{\url{https://commons.wikimedia.org/wiki/File:Portal,_Pitam_Niwas_Chowk,_City_Palace,_Jaipur,_20191218_1000_9059.jpg}}\\
\includegraphics[height=2em]{images/pont_orig.jpg}
  Daniel Vorndran / DXR (Wikimedia Commons),
  CC BY-SA 3.0\footnote{\url{https://commons.wikimedia.org/wiki/File:Pont_de_Bir-Hakeim_and_view_on_the_16th_Arrondissement_of_Paris_140124_1.jpg}}\\
\includegraphics[height=2em]{images/birds_orig.jpg}
  Prasan Shrestha (Wikimedia Commons),
  CC-BY-SA\footnote{\url{https://commons.wikimedia.org/wiki/File:Phalacrocorax_carbo,_Egretta_garzetta_and_Mareca_strepera_in_Taudha_Lake.jpg}}\\
\includegraphics[height=2em]{images/aurora_orig.jpg}
  Olga Maksimova (Wikimedia Commons),
  CC BY 4.0\footnote{\url{https://commons.wikimedia.org/wiki/File:\%D0\%9F\%D0\%BE\%D0\%BB\%D1\%8F\%D1\%80\%D0\%BD\%D0\%BE\%D0\%B5_\%D1\%81\%D0\%B8\%D1\%8F\%D0\%BD\%D0\%B8\%D0\%B5_\%D0\%BD\%D0\%B0\%D0\%B4_\%D0\%B7\%D0\%B0\%D0\%B1\%D1\%80\%D0\%BE\%D1\%88\%D0\%B5\%D0\%BD\%D0\%BD\%D1\%8B\%D0\%BC_\%D0\%B7\%D0\%B4\%D0\%B0\%D0\%BD\%D0\%B8\%D0\%B5\%D0\%BC_\%D0\%BC\%D0\%B5\%D1\%82\%D0\%B5\%D0\%BE\%D1\%81\%D1\%82\%D0\%B0\%D0\%BD\%D1\%86\%D0\%B8\%D0\%B5\%D0\%B9.jpg}}\\
\vspace{0.3em}
All other images in this article are results produced by the author with the described algorithm.
}


\bibliographystyle{siam}
\bibliography{article}

\end{document}